\documentclass[runningheads]{llncs}

\usepackage{graphicx}
\usepackage{graphicx}  
\usepackage{float}
\usepackage{hyperref}
\usepackage{xcolor}
\usepackage{multirow}
\usepackage{tabularx}
\usepackage{amsmath,amsfonts}
\usepackage{comment}
\usepackage{caption}
\usepackage{subcaption}
\usepackage{wrapfig}
\usepackage{booktabs}

\newtheorem{defn}{Definition}
\begin{document}
\title{Goal-Oriented Next Best Activity Recommendation using Reinforcement Learning }
\titlerunning{Goal Oriented Next Best Activity Using RL}
%
%



\author{Prerna Agarwal\thanks{Authors have equal contribution}\inst{1}
\and Avani Gupta\small{*}\inst{2} 
\and Renuka Sindhgatta\inst{1}
\and Sampath Dechu \inst{1}}
%
%
\institute{
    IBM Research AI, India \\
    \email{\{preragar, sampath.dechu\}@in.ibm.com, renuka.sr@ibm.com}
    \and
    IIIT Hyderabad, India \\
    \email{avani.gupta@research.iiit.ac.in}
}

\authorrunning{P. Agarwal et al.}

%
%
\maketitle              
\begin{abstract}

Recommending a sequence of activities for an ongoing case
requires that the recommendations conform to the underlying business
process and meet the performance goal of either completion time or process outcome. Existing work on next activity prediction can predict the future activity but cannot provide guarantees of the prediction being conformant or meeting the goal. Hence, we propose a goal-oriented next best activity recommendation. Our proposed framework uses a deep learning model to predict the next best activity and an estimated value of a goal given the activity. A reinforcement learning method explores the sequence of activities based on the estimates likely to meet one or more goals. We further address a real-world problem of multiple goals
by introducing an additional reward function to balance the outcome of a recommended activity and satisfy the goal. We demonstrate the effectiveness of the proposed method on four real-world datasets with different characteristics. The results show that the recommendations from our proposed approach outperform in goal satisfaction and conformance compared to the existing state-of-the-art next best activity recommendation techniques.
\end{abstract}
\section{Introduction}
\vspace{-0.2cm}
There has been a growing interest in the research community in providing decision support with studies focusing on prescriptive process monitoring and naturally extending predictive process monitoring (PPM). The focus of research in PPM has led to investigating many deep learning approaches having improved accuracy as compared to the classical machine learning algorithms. ~\cite{transformers2020,camargo2020,tax2017,taymouri2020}. However, while PPM is capable of identifying the likelihood that an ongoing case can get delayed, execute a non-conformant event, or may have a long execution trace, it cannot suggest or prescribe the optimal execution of a case that can avoid an undesirable outcome. 

Consequently, prescriptive process monitoring has addressed some of the problems. One approach is to prescribe an intervention by triggering an alarm when the probability that a case will lead to an undesired outcome is above a threshold~\cite{metzgeralarm2020,tinemaalarm1028}. An alternative approach is to recommend actions~\cite{presconfroti2013,presGroger14}. Nonetheless, there is limited work on recommending actions in the context of business goals ~\cite{adityarl2021,weinzierl2020prescriptive}. Goals are often defined for any real world business process based on Key Performance Indicators (KPI) such as cost, quality, or time.
Additionally, these goals can be contradictory. Consider an example of the loan application process - completing the loan sequence with limited actions (corresponding to activities) and approving a loan application within a stipulated time may not meet the quality requirements as the loan may be approved without checking an applicant's credit score. Hence, the goals of completion time and quality are contradictory. Moreover, the sequence of activities recommended must also conform to certain policies or regulations~\cite{conformance2018}. 

Reinforcement learning (RL) provides a framework to optimize sequences of decisions for long-term outcomes. For example, given an ongoing case, if a decision needs to be made on the next activity, each choice can affect the KPI (such as the completion time) and eventually the goal (i.e., did it complete within a stipulated time). An RL algorithm can choose the next activity (known as action) according to its policy based on the KPI estimation and receive immediate outcome (known as reward).
Hence, we propose using RL to recommend goal-oriented next best activity that conform to certain specified constraints. 

There are various challenges to use RL in the context of a goal-oriented and conformant activity sequence prediction. First, the action space varies as there are constraints on what activities can or cannot occur. 
Next, the RL model tends to choose a path that satisfies the goal without considering the outcome. It could be possible that, for instance, the model has to choose a path with the least duration and the least duration path is an undesired outcome (e.g., rejecting a loan application). In such scenarios, the model would recommend an undesirable outcome for all cases, i.e., rejecting all the loan applications.

Therefore, considering the challenges discussed above, in this work, we aim to address the following research question: \textit{Given an ongoing case, can we learn from historical executions to recommend the next best activity (and the sequence) that is goal-oriented and conformant}. Our specific contributions are as follows:
\vspace{-0.1cm}
\begin{itemize}
    \item We propose an RL based framework that recommends the next best activity (and remaining sequence of activities known as suffix) 
    given a partially completed case\footnote{The codebase is available at: \url{https://github.com/avani17101/go-nba}}. Our framework supports: (1) Deep learning model to predict the estimated value of KPI that forms the basis of RL exploration. 
    (2) Deep RL on-policy method to recommend next activity that is conformant to the process model. 
    (3) Balancing reward mechanism to mitigate the possible bias of the recommended sequence outcome.
    \item We demonstrate the effectiveness of our proposed framework by conducting experiments and comparing our work with existing prescriptive process recommendations of activity sequences.
\end{itemize}
The paper is organized as follows. A brief overview of previous studies on recommending next activity and use of RL in business process monitoring is presented (Section~\ref{sec:relwork}) followed by an introduction to the concepts in Section~\ref{sec:prelim}. The details of our approach to build the RL based framework is presented in Section~\ref{sec:approach}. The evaluation of the approach and the discussions on real-world event logs is presented in Section~\ref{sec:eval}. Finally, we summarise the contributions of our work and outline future work in Section~\ref{sec:conclusion}.

\section{Related Work} \label{sec:relwork}
\vspace{-0.2cm}
We position our work with respect to the existing work, categorized into the following three topics:
\vspace{-0.3cm}
\paragraph{Predictive process monitoring:} A large body of existing work focuses on predicting the next activity and the most likely remaining sequence of activities (suffix). Most of the recent work here relies on Recurrent architectures such as LSTMs for predicting the next activity and remaining time of cases \cite{camargo2020,mm-pred,tax2017}. Additionally, the use of transformers~\cite{transformers2020} has been explored that is efficient to train and can handle large sequences. Further, a generative adversarial architecture (GAN) has been proposed, where the generator produces sequences from random noise and the discriminator differentiates between real sequences and the ones created by the generator in~\cite{taymouri2021,taymouri2020}. The objective is to predict the sequence of activities at any point in the case and detect the likelihood of an undesired outcome, but they do not 
provide recommendation that can prevent an undesired outcome.
\paragraph{Intervention-based recommendations:}Early work on prescriptive process monitoring proposed a mechanism for generating alarms that could lead to interventions and mitigate or prevent undesired outcomes~\cite{tinemaalarm1028}. In their work, Teinemaa et al. propose an alarm system comprising of two components: 1) a probabilistic classifier that estimates the likelihood of an undesired outcome, and 2) an alarming threshold component that alarms when the likelihood of the undesired outcome is at least a threshold. The optimal alarming threshold is based on the cost of an intervention. Metzger et al.~\cite{triggering-2020} use an RL algorithm to automatically learn when to trigger process adaptations based on the predictions.  Recent work proposed a framework that recommends when to apply an intervention (treatment) to an ongoing case to decrease its cycle time by building causal models~\cite{prescriptiveirene21}. Orthogonal random forests trained on historical traces is used to estimate the effect of a treatment (or intervention) on the reduction in cycle time, given the current state of a case. In these approaches, the recommendation of the intervention or treatment is binary and does not recommend an optimal next best activity or suffix.
\paragraph{Suffix recommendations:}
Recent work on prescriptive process monitoring proposes the next best event recommendation that is optimized for a given KPI~\cite{weinzierl2020prescriptive}. A DL model is trained to predict next activities and the KPI values. For an ongoing case, the suffix and the KPI values are predicted. A set of $k$ nearest neighbors of the predicted suffix is chosen from historical data as candidate suffixes. They are then passed to a business process simulation model to reduce the risk of non-conformant activities and recommend the best suffix. Non-conformant activities are simply discarded in candidate suffixes. But, when all candidate suffixes are non-conformant, the predicted suffix is returned as the best action. In this particular case, the recommendations would be non-conformant. Hence, this work also does not guarantee conformant recommendations. Moreover, a pre-defined $k$ neighbors in the search space is needed to select the suffix meeting the goal. Therefore, we distinguish our approach by its ability to support multiple goals (or KPIs) and explore a search space of optimal paths while providing conformant recommendations using RL. An approach using Memory augmented neural networks (MANN) recommends suffixes by learning from historical data labeled as meeting or deviating from the goal~\cite{adityaMANN2021}. In a later work, Khan et al.~\cite{adityarl2021} use an RL algorithm to recommend suffixes that meet a goal. While the authors use RL to recommend the best actions, they do not consider  
multiple conflicting goals with the additional constraint of ensuring conformant execution paths.

\section{Preliminaries} \label{sec:prelim}
\label{preli}
In this section, we provide the background notions useful in the rest of the paper. 
\begin{defn}
\textbf{Event log}. An event log is a set of $m$ traces $E = \{\sigma_1,\sigma_2,...,\sigma_m\}$.
\end{defn}

\begin{defn}
\textbf{Trace}. A trace $\sigma = \langle e_1,e_2, \dots, e_n \rangle$ is an ordered sequence of $n > 0$ events. A partial sequence of length $k \leq n$ would consist of the first k elements of the sequence. Each trace have a unique identifier $trace\_{id}$. 
\end{defn}

\begin{defn}
\textbf{Event}. An event $e$ is a tuple of trace id $trace\_{id}$, activity $a$ and timestamp $ts$.
$$e = (trace\_{id}, a, ts)$$
\end{defn}

\begin{defn}
\textbf{Key Performance Indicator (KPI)}. The performance of trace $\sigma$ can be measured on multiple dimensions such as time, cost, or quality. The value $v$ obtained for a trace $\sigma$ at an activity $a$ for a performance indicator $pi$ is termed as Key Performance Indicator. $KPI(\sigma,a,pi) = v_a^{pi}$ 
\end{defn}
\begin{defn}
\textbf{Goal value and Goal}. A goal value $G$ for an trace $\sigma$ is defined as the cumulative value of a KPI. Hence, $G(\sigma,pi) = \sum_{a \in \sigma} KPI(\sigma,a,pi)$. A Goal is defined as a satisfaction threshold $\omega$ on the goal value that each event sequence is expected to meet. If $G(\sigma) \leq \omega$ (for minimization problem) or $G(\sigma,pi) > \omega$ (for maximization problem), the trace $\sigma$ is said to satisfy the goal (GS) otherwise it is said to violate the goal (GV). 
\end{defn}

\begin{defn}
\textbf{Process Outcome}. The outcome of a trace $\sigma$ is defined as the last activity in the sequence i.e., outcome = $Last_a(\sigma)$. The set of all such possible outcomes defines outcome set $O$. 
\end{defn}


\section{Approach} \label{sec:approach}
\begin{figure}[t]
\centering
\includegraphics[width=0.9\linewidth, height=5cm]{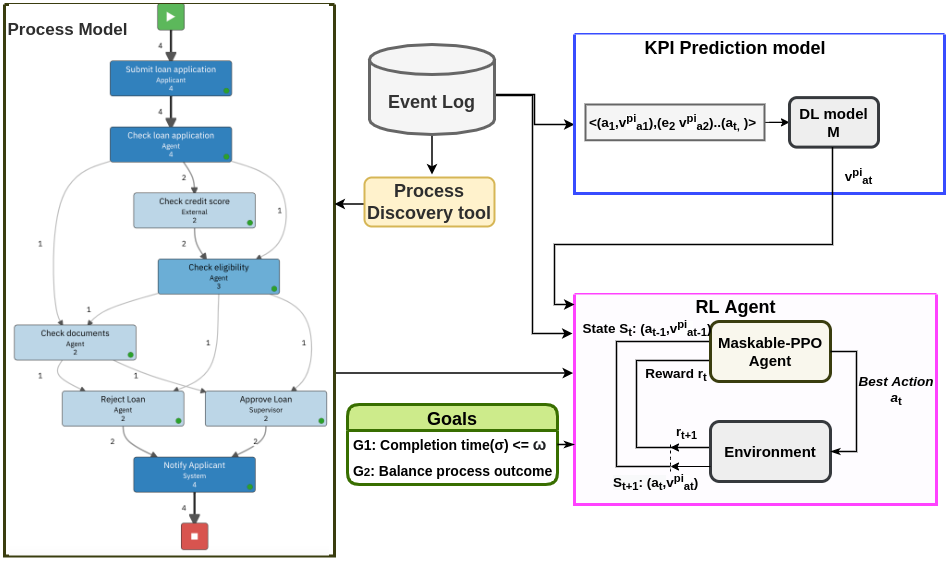}
\caption{Proposed Framework}
\label{fig:arch-diagram}
\end{figure}

\vspace{-0.2cm}
The architecture of our proposed method is shown in Figure \ref{fig:arch-diagram}. First, we take a historical sequence of events as our input event log $E$. Given E, we use the IBM Process Mining tool\footnote{\url{http://ibm.com/cloud/cloud-pak-for-business-automation/process-mining}} to discover a \emph{directly-follows-graph}(DFG) that enables us to verify sequences or paths that are conformant to a reference process model. Next, a deep learning-based model $M$ is trained on activity sequences in $E$ to predict KPI for a goal $G$. The RL agent takes as input the predicted KPI, partially completed activity sequences, the goal $G$, and the discovered DFG to learn only the activity sequences that satisfy the goals while conforming to the DFG. We next describe the technical details of each component in detail.

\subsection{Business Process Directly-Follows Graph} 
The event log $E$ contains event sequences generated through the execution of the business process instances (referred to as cases). 
A directly follows graph (DFG) representing an approximate business process is discovered using the existing \emph{Process discovery} algorithms \cite{leemans2019,dfgsdm2021} (Figure~\ref{fig:arch-diagram}). 
The primary requirement is to ensure that the predicted activity sequences conform to the discovered business process DFG.  
The absence of DFG could lead to recommending non-conformant sequences. This is highly undesirable as non-conformance can lead to violation of regulations and legal obligations. Hence, we use the DFG to guide the activity sequence prediction model and learn only the sequences that conform to the business process. 

\subsection{Key Performance Indicator Prediction Model}
Key performance indicators (KPI) such as cost, time, or quality must be measured for each event in a trace. 
Predicting KPI values after executing each activity of a partial sequence helps in estimating their corresponding goal value and checking whether unfolding the execution of the entire sequence of activities by choosing that specific activity would violate the goal or not. The KPI prediction model is used to explore different activities and choose the ones that will satisfy the goal. We need to consider that the chosen activity may not satisfy the goal at the current point in time but could eventually satisfy it when the case ends. As depicted in Figure~\ref{fig:arch-diagram}, we train the KPI prediction model with the input as: (1) the partially completed activity sequence and (2) the activity chosen by the RL agent (discussed in Section~\ref{rlagent}). The objective is to predict the KPI given the partially completed activity sequence with their KPI values and the next chosen activity. Existing KPI prediction models only consider the partially completed activity sequence as input. As we require to explore all conformant paths, our model considers the partial sequence and the next likely activity. 

Any sequence prediction deep learning-based model such as LSTM, CNN, Transformers can be used for this purpose. We choose GAN-LSTM \cite{taymouri2020} to build the KPI prediction model as it provides higher accuracy as compared to other models. The input to the model is the partially completed activity sequence $\langle (a_1, v_{a_1}^{pi}), (a_2, v_{a_2}^{pi}), ..., (a_t) \rangle $ where $a_i$ is the activity and $v_{a_i}^{pi}$ is the KPI value at activity $a_i$. The output $v_{a_t}^{pi}$, is the predicted KPI for the activity $a_t$.

\subsection{Goal-Oriented Model using Reinforcement Learning}
To recommend the next activity that conforms to the business process and is goal-oriented, we use a reinforcement learning based-method to explore all the activities that conform to the DFG and determine which activity(s) would lead the sequence towards the fulfillment of the goal. To achieve this, we have carefully crafted the learning problem specifically for goal-oriented learning using the event log guided by the DFG and KPI prediction model.
\vspace{-0.3cm}
\subsubsection{Learning Problem}\label{Learning Problem}
We formalize the learning problem of next activity recommendation by defining action space $A$, state space $S$, and reward function $r$. Each event in DFG and its corresponding KPI value becomes the state. The action space would vary at every state according to the possible actions obtained from business process DFG.  For example: consider a DFG where we are at an activity \textit{P} and there are three parallel paths leading to the next activity \textit{Q, R}, or \textit{S} respectively, then there can be three possible actions in $A$. Whereas after activity \textit{S}, the only possible next activity is \textit{T} and hence the possible action is different.
This is not the case in a general RL setting, as the action space remains constant at all states. However, in our learning problem, it will vary at each state. We use the \emph{Masking} technique in RL to deal with this challenge. The details are discussed in Section~\ref{rlagent}.

The core part of formalizing the learning problem is to define the reward function $r$. Since our aim is to recommend the next best activity (and sequence) that will satisfy the goal, we predict the KPI value at each state and calculate the goal value $G(\sigma)$ when the end of the trace $\langle EOS \rangle$ is reached. We then provide the reward at $\langle EOS \rangle$ based on the goal value. For all other states, we provide a constant reward that is $0$ for our purpose.
The goal value is then checked against the goal satisfaction criteria ($\omega$) to determine a goal violation (GV) or goal satisfaction (GS). 
The reward function is shown in Table \ref{tab:reward-function}.
\begin{table}[t]
    \centering
    \caption{Reward function $r$ based on KPI prediction}
    \begin{tabular}{c|c}
         \textbf{Goal Satisfaction (GS)}&  \textbf{Goal Violation (GV)}\\
         \hline
         \hline
         \large $\frac{2 * abs(G(\sigma)-\omega)}{\omega}$ & \large $\frac{-2 * abs(G(\sigma)-\omega)}{\omega}$
    \end{tabular}
    \label{tab:reward-function}
\end{table}
Note that the magnitude of reward is directly proportional to the deviation of $G(\sigma)$ from $\omega$.
When its minimization problem, Goal Criteria becomes $GS: G(\sigma) \leq \omega$ (as defined in Section \ref{preli}), reward for GS is $2*\frac{G(\sigma)-\omega}{\omega}$, which ensures more positive reward for smaller $G(\sigma)$. For GV, the reward $- 2*\frac{G(\sigma)-\omega}{\omega}$ ensures more negative reward for larger $G(\sigma)$. The reward function defined above remains same for the maximization problem as well i.e., $GS: G(\sigma) > \omega$. \vspace{-0.2cm}
\subsubsection{RL Agent} \label{rlagent}
The action space $A$ for the RL agent is the set of available activities derived from DFG. The state space $S$ at time $t$ consists of a tuple of previous predicted activity $a_{t-1}$ and it's corresponding KPI value $v^{pi}_{a_{t-1}}$. As shown in Figure \ref{fig:arch-diagram}, at time $t$, the agent is in state $s_t \in S$ i.e., $(a_{t-1},v^{pi}_{a_{t-1}})$. It samples an action $a_{t} \in A$. Then, given $a_{t}$, the KPI prediction model predicts the corresponding KPI value i.e., $v^{pi}_{a_{t}}$. The RL agent gets reward $r_t$ and moves to the next state $s_{t+1}$. The cumulative reward is calculated when it predicts end of trace $\langle EOS \rangle$.
We define the available actions as per DFG as valid and all other actions as invalid. We use the state-of-the-art on-policy Proximal Policy Optimization (PPO) algorithm~\cite{schulman2017proximal} in our method.

\textbf{PPO} ~\cite{schulman2017proximal} has been widely used in the RL community due to its ease of implementation, superior performance, and less hyper-parameter tuning requirement.
We use the \emph{Advantage Actor-Critic} version of PPO. A policy $\pi: s \rightarrow a $ is estimated based on the interaction of agent with environment. In state $s_t$, the Actor takes the action $a_t$ according to the current policy $\pi_{\theta}(s_t, a_t)$ resulting in a trajectory $T$ and reward $r_t$. $\theta$ represents neural networks parameters.
\begin{equation} 
T = \left(s_{t}, a_{t}, \pi_{\theta}(s_t, a_t), r_{t}\right): 1 \leq t \leq M 
\label{T_eqn}
\end{equation}
where, M is the total number of time-steps the model takes.
The $r_t$ received is then evaluated by Critic which outputs the value of state $v(s_t)$. The advantage function $A_{t}$ gives information about the extra reward an agent can get by taking the action $a_t$. $A_t$ is calculated as the difference between Q-value $Q^\pi_{\theta}(s_t, a_t)$ and $v(s_t)$. 
$Q^\pi_{\theta}(s_t, a_t)$ is the sum of current reward and estimated future rewards discounted by $\gamma$ obtained by taking action $a_t$ at time t according to $\pi_{\theta}(s_t, a_t)$. 
\begin{equation} 
Q^\pi_{\theta}(s_t, a_t) = r_{t}+\gamma r_{l+1}+\gamma^{2} r_{t+2} \cdots+\gamma^{N} r_{t+N}
\end{equation}
where, $N$ is number of time-steps for which the future reward is estimated. The loss functions are used in a similar manner as described in \cite{schulman2017proximal}.

\textbf{Variable action space}
There are two ways to handle variable action space: (1) Giving a large negative reward for invalid actions, (2) \textit{Action masking}: 
Masking out the actions as valid or invalid in order to sample action only from valid actions. We experiment with both approaches. 
For Action masking, we experiment with the maskable-PPO technique proposed in \cite{tang2020implementing}.

\textbf{Maskable-PPO}
Action masking-based PPO~\cite{tang2020implementing} is implemented by modifying the original PPO implementation as follows:
(1) Trajectory $T$ (Eq. \ref{T_eqn}) is collected only for valid actions. 
(2) Only valid actions are used to  calculate stochastic descent.
Softmax value is used at the end of the output layer. For re-normalization, softmax value is computed only for valid actions. The probability of invalid actions is made zero, and the action probabilities are re-normalized.

\textbf{Comparison with other RL algorithms} 
We compare the on-policy PPO-based proposed method with the state-of-the-art off-policy Deep-Q-Network (DQN) algorithm~\cite{mnih2013playing}. In contrast to on-policy, the agent takes random actions to determine $Q$ values in off-policy (without a policy). 
To handle variable action space, we experiment with both PPO and DQN by giving a large negative reward for invalid actions.  
The comparison of the results are discussed in Section~\ref{results}. 
\vspace{-0.6cm}
\subsubsection{Trade-off}
The current reward function can become biased towards the activities that consistently satisfy the goals. Consider the example of a loan approval process with two possible process outcomes: \emph{loan approved} and \emph{loan rejected} with the goal of the completion time ($G(\sigma, time) \leq \omega$). It may be possible that the time for processing an application with the outcome as \emph{loan rejected} is less as compared to the outcome \emph{loan approved}. The existing reward function may learn to reject the loan, leading to high goal satisfaction but with a high bias for the outcome of the process, which is undesirable. 

We address this limitation by considering an additional secondary goal that allows us to match the process outcome of the generated sequences with the ground truth distribution of process outcome. We provide an example of one such secondary goal, but our approach is flexible to accommodate other definitions of secondary goals (e.g., equal distribution of tasks to resources). Hence, an additional balancing reward $\delta r_i$ (Table \ref{tab:reward-function-2}) is used to achieve a trade-off that supports the primary goal based on the KPI and secondary goals. The balancing reward $\delta r_i$ tries to mimic the distribution of different process outcomes using the ground truth traces. 
For this, we provide a reward at the states which majorly contribute towards the process outcome along with the reward $r$. We hypothesize that the last $k$ activities of a trace can influence the process outcome the most. The value of $k$ could vary with different processes. Therefore, for these $k$ activities, if the action $a$ chosen by the agent is equal to the ground truth event ($gt$), we provide a positive reward of +0.5 else penalize with the reward of -0.5. 
\vspace{-1cm}
\begin{table}[H]
    \centering
    \caption{Balancing reward factor $\delta r_i$}
    \begin{tabular}{c|c}
        \textbf{a=gt}& \textbf{a!=gt}\\
        \hline
        \hline
        +0.5 & -0.5
    \end{tabular}
    \label{tab:reward-function-2}
\end{table}
\vspace{-0.8cm}
We denote $\delta r_k$ as cumulative balancing reward for last $k$ activities.
\vspace{-0.2cm}
$$ \delta r_k = \sum_{i=1}^{k} \delta r_i$$
The final reward $r'$ becomes:
$r' = r + \delta r_k$. Hence, there is a trade-off between the percentage of traces that satisfy the primary goal and the secondary goal, mitigating the skewness. 

\section{Experimental Evaluation} \label{sec:eval}
In this section, we describe the datasets used to evaluate our method under different experimental settings along with training details. We also compare our method with next best action recommendation baseline.
\subsection{Datasets}
We use 4 real-world event datasets popularly used to evaluate activity prediction tasks. The descriptive statistics of these event logs are shown in Table~\ref{tab:datasets}. The \#activity denotes the number of unique activities present in dataset, and $|\sigma|$ is the sequence length.
\begin{itemize}
    \item \textbf{Helpdesk}: An event log of a ticket management system of an Italian software company\footnote{\url{https://data.4tu.nl/articles/dataset/12675977}}. 
    \item \textbf{BPIC12W}: An event log of a loan application process of a Dutch Financial institution\footnote{\url{https://data.4tu.nl/articles/dataset/12689204}}. 
    \item \textbf{Road Traffic Fine Management Process}: An event log of a road traffic fine management system\footnote{\url{https://data.4tu.nl/articles/dataset/12683249}}.  
    \item \textbf{BPIC2019}: An event log of purchase order handling process data from a large multinational company operating from The Netherlands\footnote{\url{https://data.4tu.nl/articles/dataset/12715853}}.
\end{itemize}
Traffic Fine and BPIC2019 data contains very large number of traces, hence, we randomly sample 10\% of the data for experimentation.
\begin{table}[t]
    \centering
     \caption{Datasets used in experiments}
    \begin{tabular}{|c|c|c|c|c|c|c|c|}
    \hline
         \textbf{Dataset}&\textbf{\#$\sigma$}& \textbf{\#events} & \textbf{\#activity} & \textbf{Mean $|\sigma|$} & $\mathbf{C\% }$& $\mathbf{dur_{avg}}$\textbf{(days)}&\textbf{$\omega$(days)}\\
         \hline
         \hline
         Helpdesk & 3804 & 13710 & 10 & 3.60& 81.83  & 8.49 & 13.89\\ 
         BPIC12W & 9658 & 72413 & 7 & 7.50& 68.41  & 15.72 & 24.001\\
         Traffic Fine & 15037 & 56388 & 12 & 3.75& 83.60 & 342.41 & 607.04\\
         BPIC2019 & 5241&31736& 31& 6.05& 67.16 & 625.50 & 791.23\\
         \hline
    \end{tabular}
    \label{tab:datasets}
\end{table}

\subsection{Baselines}
To the best of our knowledge, this is the first attempt to learn a goal-oriented activity recommendation model. Hence, there is no goal-oriented learning baseline available for comparison. Therefore, we compare our work with Next best action recommendation (NBA) by Weinzierl et. al \cite{weinzierl2020prescriptive} that reduce the non-conformant predictions using DCR (Dynamic Condition Response) graph. This baseline could be compared only for the helpdesk and BPIC2019 dataset because of the unavailability of the DCR graph for the other two datasets. 
For this baseline, the end of the trace is marked when the model predicts either $ \langle EOS \rangle$ or when the trace length exceeds the maximum allowed trace length in the dataset.
We compare the percentage of recommendations that (1) satisfy the goals and (2) conform to DFG. We also compare the Damerau-Levenshtein (DL) distance between the recommended activity sequence and ground truth.

\subsection{Experimental Setup}
This section provides details of data pre-processing, KPI prediction model, and the RL agent.
\vspace{-0.45cm}
\subsubsection{Data Pre-processing}
The event log contains traces each having a unique id i.e., $trace\_{id}$. 
Similar to the GANPred model by Taymouri et al. \cite{taymouri2020}, and Tax et al.\cite{tax2017}, we denote the time to complete an activity, i.e., the activity time, as the time difference between two consecutive activities in a trace in days. Hence, for an activity $a_i$ for an event $e_i$, the activity time $at_i$ is calculated as the difference between timestamp of the previous activity of the event i.e., $t_{i-1}$ and timestamp of the current activity of the event i.e., $t_i$. 
\vspace{-0.5cm}
\subsubsection{KPI prediction model}
We encode the activities using one-hot vector encoding. The model predicts the \emph{event time} for activities. We generate partial sequences for each trace in the event log. For example, for a complete sequence consisting of a total of 5 activities $\langle a, b, c, d, e \rangle$, we could consider up to 4 partial sequences i.e., $\langle a \rangle$, $\langle a, b \rangle$, $\langle a, b, c \rangle$, $\langle a, b, c, d \rangle$. 
We then train a separate model for each sequence length. Hence, an event log containing complete sequences of length $m$ would have $m-1$ models, each catering to a partial sequence of length 1 to $m-1$. A train-test split of 80-20 is used similar to~\cite{taymouri2020}.
\vspace{-0.5cm}
\subsubsection{For RL agent}
The RL agent learns sequences from the input data. We want the RL agent to predict activity sequences that conform to the DFG. We also need to match the outcome of the sequence with the ground truth sequence outcomes in order to mitigate the skewness during learning. 
The ground truth sequences should also conform to the discovered DFG. But, often, the discovered DFG from the event log is an approximate representation and hence does reflect all the activity sequences~\cite{conformance2018}. Therefore, for fair training and comparison, we filter and use only the sequences that conform to the DFG for training RL agents. We take goal satisfaction threshold ($\omega$) as the third quartile value of case duration in the entire dataset for our experiments. The percentage of sequences in ground truth that conforms to DFG (C\%) along with $\omega$ is shown in Table \ref{tab:datasets}. We split this filtered dataset into train and test set, keeping a ratio of 65:35. We denote the average duration of a sequence by $dur_{avg}$ and report in Table \ref{tab:datasets}.
Note that the validation set is not required since the RL agent is trained in an online manner.

\subsubsection{Experimental Details}
We implemented our environment in Open-AI Gym \cite{1606.01540}. The RL agent is given the first activity of each sequence as input. It generates the complete sequence by choosing the next best action from the action space $A$, leading towards the goal satisfaction.

First, we train the KPI model. It is then used by the RL agent to obtain the predicted KPI values for each action it chooses.
The input to the KPI model is the partial sequence consisting of the first activity and subsequent activities chosen by the RL agent and the current action chosen by the RL agent to explore. The model then predicts the KPI corresponding to the current action chosen. 
Since the action space, i.e., the set of available actions at each state, varies according to the business process DFG, the exploration is based on the available actions, and the best action is chosen. PPO agent explores by sampling actions according to its stochastic policy, which keeps on changing as it learns. That means that the agent learns on the go and its experience changes the policy and hence the exploration. 
The next best action is chosen sequentially after each activity is predicted until the RL agent predicts the $\langle EOS \rangle$. Once $\langle EOS \rangle$ is predicted, the end reward is calculated along with the balancing reward (if enabled). We experiment with both scenarios and report the results with different $k$ values.

We have considered sequence completion time (cumulative event time) as goal value with event time as KPI in our experiments. We define the goal as to minimize the time taken by each sequence with $\omega$ as GS threshold (see Table \ref{tab:datasets} for $\omega$ values). To accommodate for the errors in the KPI prediction model, we relax the goal satisfaction bound with $\pm$ MAE (mean absolute error) of the KPI prediction model at each sequence length. Since the KPI prediction model is trained for each partial sequence, therefore, MAE for each partial sequence length is available. We relax the bound by the cumulative sum of MAE's until the predicted sequence length. If a sequence $\sigma$ has a predicted partial sequence of length $m$, we compute the MAE as follows: 
\vspace{-0.3cm}
$$ MAE(\sigma) = \sum_{i=1}^{m} MAE_i \\ $$
\vspace{-0.2cm}
So our new GS condition becomes, 
$$ GS: G(\sigma) \pm MAE_{\sigma} \leq \omega $$


The ground truth contains both GS and GV sequences. The aim is to recommend the sequence that leads to goal satisfaction at the end. Therefore, we measure the percentage of GV sequences in ground truth for which the RL agent recommended the sequence that satisfied the goal. It is an important metric to measure the performance of RL agent. 

The learning rate for critic is taken as 0.001, the learning rate for actor as 0.0003, PPO epsilon clip as 0.2, PPO update frequency as 32000000 time-steps and discount factor ($\gamma$) as 0.99\footnote{\url{https://github.com/nikhilbarhate99/PPO-PyTorch}}.
The KPI prediction model is trained over each sequence length for 25 epochs. It is trained on 1-2 CPU's with 2 GB memory each. The RL agent is trained on one 1040 GTX GPU and 10 CPU's each taking a memory of 2GB. RL agent is trained for 300-500 epochs. 
The RL agent is able to achieve near-optimal performance in 20 epochs. 

\begin{table*}[t]\small
    \centering
    \caption{Results on all Datasets}
    \begin{tabular}{|p{1.6cm}|p{1.1cm}|p{1.35cm}|p{2.2cm}|p{1.5cm}|p{1.8cm}|c|c|c|}
 \hline
  \textbf{Dataset} &                            \textbf{GS (GT)}\%             & $\mathbf{MAE_{KPI}}$ \textbf{(days)}                    &   \textbf{Reward type   }                        &  $\mathbf{GS_{pred}}$(\%) &  \textbf{GV turned GS}\% & $\mathbf{Acc_{1} \%}$&  $\mathbf{Acc_{2} \%}$ & $\mathbf{Acc_{3} \%}$ \\
 \hline
\multirow{4}{4em}{Helpdesk} &  \multirow{4}{4em}{75.95}       & \multirow{4}{4em}{0.9716}& $r$                   &\textbf{90.87} &\textbf{90.83} &  100 &    60.70 & 46.49 \\
                            &                                 &                          &   $r'$ with $k=1$     &     88.51   &  90.71        &        100  &                         60.56 &                          47.07\\
                            &                                 &                          &   $r'$ with $k=2$     &      90.01 &            90.63       &  100 &    \textbf{61.41} &                          47.77 \\
                            &                                 &                          &   $r'$ with $k=3$     &      89.14 &            90.11 &                100 &                         61.23 &                          \textbf{48.05} \\
\hline
\multirow{4}{4em}{BPI1C2W}   &  \multirow{4}{4em}{76.41}       & \multirow{4}{4em}{0.3052}& $r$        &       \textbf{99.90} &            \textbf{99.90}        &        36.38 &                         30.18 &  27.42 \\
                            &                             &                          &    $r'$ with $k=1$      &        99.98      &       99.87 &                 \textbf{37.90} &                         31.80     &                       29.43\\
                            &                                 &                          &    $r'$ with $k=2$    &        99.94      &      99.93 &              36.65 &                  \textbf{36.04} &                          28.54 \\
                            &                                 &                          &    $r'$ with $k=3$    &       90.87     &       90.86 &              36.46 &                         31.43 &                          \textbf{35.94} \\
\hline
\multirow{4}{4em}{$Traffic_{ss}$}& \multirow{4}{4em}{72.67}  & \multirow{4}{4em}{39.8247}&  $r$   &      \textbf{99.99} &           \textbf{99.98} &     30.84  & 32.07 & 28.54\\
                             &                               &                           &    $r'$ with $k=1$       &     99.97 &           99.97 &                \textbf{31.15}  &                  32.16 &                          29.40 \\
                             &                               &                           &    $r'$ with $k=2$       &     99.94 &           99.93 &                  28.11 &      30.78 &                          28.59\\
                             &                               &                           &    $r'$ with $k=3$       &     99.12 &           99.11 &                30.92 &                         \textbf{32.26} &                          \textbf{29.49}\\
        \hline
\multirow{4}{4em}{$BPIC2019$}& \multirow{4}{4em}{80.32}  & \multirow{4}{4em}{39.8247}&  $r$   &  \textbf{89.78}    &   \textbf{89.86}     &   35.53   &32.14  & 28.78\\
                             &                               &                           &    $r'$ with $k=1$       &    88.52  &           89.23 &                \textbf{35.89}  &                  31.78&                          28.38 \\
                             &                               &                           &    $r'$ with $k=2$       &  88.21  &       89.23  &              35.19   &  \textbf{32.67}    &          28.38           \\
                             &                               &                           &    $r'$ with $k=3$       &  86.48  &   86.51  &  34.22          & 31.93                      &       \textbf{30.63}                   \\
        \hline
    \end{tabular}
    \label{tab:results}
\end{table*}
\begin{figure}
\centering
\includegraphics[scale=0.45]{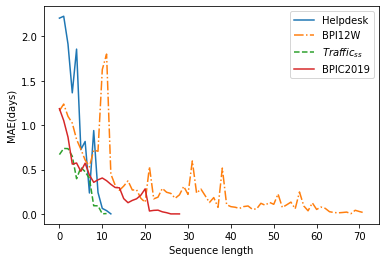}
\caption{MAE of KPI prediction model for different sequence lengths. The MAE for $Traffic_{ss}$ and BPIC2019 is factored by 0.01 to fit in the plot}
\label{fig:MAE}
\end{figure}
\vspace{-1cm}
\subsection{Results and Discussion} \label{results}
The results of our proposed method are shown in Table \ref{tab:results}. The column GS (GT) \% shows the percentage of sequences in the Ground Truth (GT) that satisfy the goal in the test set. We report the average MAE of KPI prediction model (in days) on overall sequence length in column $MAE_{KPI}$. The MAE is high for $Traffic_{ss}$ and BPIC2019 due to high average completion time of sequences in these datasets (see Table \ref{tab:datasets}). 
The MAE (as shown in Figure \ref{fig:MAE}) for different datasets on various sequence lengths decreases as the sequence length increases. The \textit{Reward type} column denotes the type of the reward being used i.e., $r$ (goal satisfaction) or $r'$ (goal satisfaction with balancing reward). The column \textit{$GS_{pred}$} shows the percentage of sequences recommended by our model that satisfies the goal. The column \textit{GV turned GS} shows the percentage of sequences in GT which were violating the goal (GV) for which our model is able to recommend the sequence that satisfies the goal (GS). 
We report the Accuracy $Acc_{k}$ for the last $k$ events as well. Computing accuracy on the last $k$ events enables comparison of the process outcome of the recommended sequences with the actual sequences.
We report the results for $k = \{1,2,3\}$. All the results are reported as the average over 100 test episodes.

\begin{figure}[htb]
     \centering
     \begin{subfigure}[b]{0.47\textwidth}
         \centering
         \includegraphics[width=\textwidth]{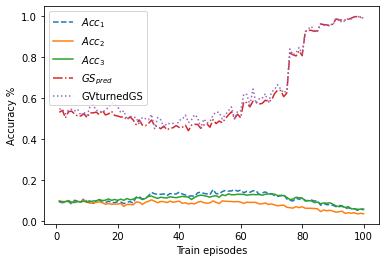}
         \caption{Trade-off accuracy between goal satisfaction and process outcome for Helpdesk}
         \label{fig:tradeoff}
     \end{subfigure}
     \begin{subfigure}[b]{0.47\textwidth}
         \centering
         \includegraphics[width=\textwidth]{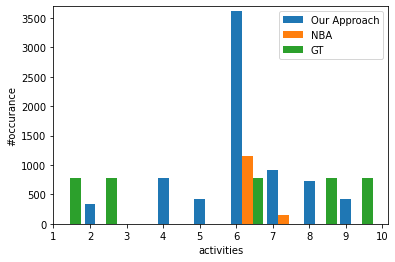}
         \caption{Comparison of last ($k$=1) recommended activity distribution on Helpdesk}
         \label{fig:distl}
     \end{subfigure}
        \caption{Trade-off accuracy and last $k$ activity distribution on Helpdesk}
\end{figure}

As shown in Table \ref{tab:results}, $GS_{pred}$\% is highest with reward type $r$ for each dataset compared to other reward types. This is because of the trade-off between the primary goal and secondary goal, as discussed in Section 4.3. The trade-off is shown in Figure \ref{fig:tradeoff}, where, as the GS, and GV turned GS increases (indicating the primary goal), the $Acc_k$ decreases (representing the secondary goal). As observed, the $GS_{pred}$\% and GV turned GS\% (primary goal) drops slightly for reward types $r'$ but $Acc_k$ increases (secondary goal). Thus, the optimal $k$ value can be chosen depending on this trade-off. The high GV turned GS\% for all datasets demonstrates the efficacy of our proposed approach and modeling of reward function to recommend the sequences that satisfy the goals for a high number of GV sequences. This is because the crux of our proposed method lies in recommending the goal satisfying sequences for GV. The $Acc_{1}$ on the helpdesk always comes out to be 100\% because the BP-DFG graph of helpdesk has just one event before $\langle EOS \rangle$.

Table \ref{tab:baselines} shows the comparison of our proposed method with NBA baseline~\cite{weinzierl2020prescriptive}.
The conformance percentage $C$ for the baselines is not 100\%, which means that the recommendations by the baseline do not guarantee conformance to the process model. In contrast, our proposed method always recommends sequences that conform to the DFG. The $GS_{pred}$\% for the baseline is less due to the lack of modeling goal information. On the other hand, our proposed method performs lesser than baseline for $Acc_k$. The baseline is trained to select the nearest suffix and hence the activities from the GT, whereas the aim of our approach is to recommend the activity that will eventually satisfy the goal and, therefore, can differ from GT. 

\begin{figure}
     \centering
     \begin{subfigure}[b]{0.47\textwidth}
         \centering
         \includegraphics[width=\textwidth]{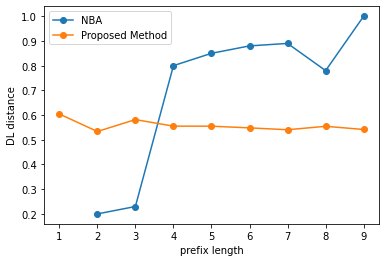}
         \caption{Helpdesk}
         \label{fig:Helpdeskdl}
     \end{subfigure}
     \begin{subfigure}[b]{0.47\textwidth}
         \centering
         \includegraphics[width=\textwidth]{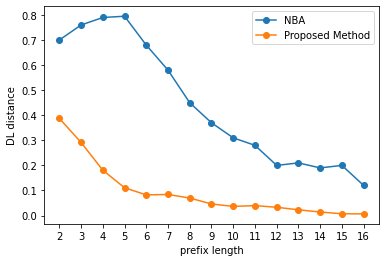}
         \caption{BPIC2019}
         \label{fig:bpic2019dl}
     \end{subfigure}
        \caption{Comparison of DL distance from GT on helpdesk and BPIC2019 datasets}
        \label{fig:dl-distance}
\end{figure}

Further, we also compare the Damerau–Levenshtein (DL) distance between the baseline (for $k$NN value as 15 as required by NBA) with our proposed approach. As shown in Figure \ref{fig:dl-distance}, the DL distance of our proposed approach for both datasets is much lower than the baseline. This shows that the exploration done by our RL agent is not too far from the GT and yet recommends the sequence of activities that satisfy the goals. 
\begin{table*}[t]\small
    \centering
    \caption{Comparison with NBA baseline}
    \begin{tabular}{|p{1.6cm}|c|c|c|c|c|c|c|}
 \hline
 \textbf{Dataset}           &\textbf{Method}&$\mathbf{C \%}$ & $\mathbf{GS_{pred} \%}$  &  $\mathbf{Acc_{1} \%}$&  $\mathbf{Acc_{2} \%}$ & $\mathbf{Acc_{3} \%}$\\
 \hline
\midrule
\multirow{2}{3em}{Helpdesk} 
                            &         NBA~\cite{weinzierl2020prescriptive} &       84.63 &  82.31 & 70.23 & \textbf{64.71} & \textbf{63.00} \\
                            &        Proposed Approach &          \textbf{100} &   \textbf{89.14} & \textbf{100} & 61.23 & 48.05\\
\hline                            
\midrule
\multirow{2}{3em}{BPIC2019}
                            &         NBA~\cite{weinzierl2020prescriptive} &       50.74 &  19.88 & \textbf{57.40} & \textbf{50.99} & \textbf{48.91} \\
                            &        Proposed Approach &          \textbf{100} &   \textbf{86.48} & 34.22 & 31.93 & 30.63\\
 \hline
    \end{tabular}
    \label{tab:baselines}
\end{table*}


The results show that our proposed algorithm is able to recommend goal satisfying sequences on an average for 93\% of GV sequences. Further, in our experiments, goal satisfaction and balancing outcome were the two conflicting goals that our method is able to handle. 
To compare how close our method is able to mimic the process outcome distribution of GT and the baseline, we plot the distribution of the last $k$=1 activity of GT, baseline, and our proposed method as shown in Figure \ref{fig:distl}. The distribution is averaged over the number of activities in the test set of the baseline for a fair comparison. 
The baseline recommends just a few activities from GT, whereas our approach not only recommends all activities from GT but also discovers other possible activities. This is achieved with the help of the balancing reward. 
\begin{table*}[t]\small
    \centering
    \caption{Comparison of different RL algorithms for $k=3$ on Helpdesk}
    \begin{tabular}{|c|c|c|c|c|c|}
 \hline
 \textbf{Method} & $\mathbf{C \%}$ & $\mathbf{GS_{pred} \%}$  &  $\mathbf{Acc_{1} \%}$&  $\mathbf{Acc_{2} \%}$ & $\mathbf{Acc_{3} \%}$\\
 \hline
\midrule
$PPO_{neg}$ &       95.58 &  10.603 & 29.32 & 24.50 & 21.56\\
$DQN_{neg}$ & 82.43 &   4.65 & 3.37 & 3.29 & 3.29\\
Maskable-PPO &  \textbf{100} &   \textbf{89.14} & \textbf{100} &\textbf{61.23} & \textbf{48.05} \\
 \hline
    \end{tabular}
    \label{tab:dqn}
\end{table*}

The comparison between PPO and DQN with large negative reward, i.e., $PPO_{neg}$ and $DQN_{neg}$ along with Maskable-PPO is shown in Table \ref{tab:dqn}. The large negative reward value is set to $-4$ for the experiments. PPO being an on-policy method, performs way better than DQN because of its capability to learn stochastic policies and efficiently handle complex Q-functions.
Maskable-PPO outperforms the large negative reward approach because a large negative reward hinders the exploration of the RL agent by shifting the aim to avoid non-conformant paths rather than focusing on choosing goal-satisfying paths.


\section{Conclusion and Future Work} \label{sec:conclusion}
In this paper, we proposed a reinforcement learning (RL) based method to predict the next best activity (and sequence) that satisfies specific goals and conforms to certain constraints. We motivate the need for such a framework in the application context of real-world business process executions. We further propose a method to handle conflicting goals.  Our work is the first to build a goal-oriented next best activity (and sequence) recommendation model to the best of our knowledge. We would extend the current framework to include multiple goals as a part of future work. We also plan to incorporate the effect of data attributes present in the logs on recommendations. Further, we plan to incorporate our work in a decision recommendation platform that recommends next best actions to the knowledge workers based on goal satisfiability.

\bibliographystyle{splncs04}
\bibliography{ref}
\end{document}